\documentclass[10pt,twocolumn,letterpaper]{article}

\usepackage[pagenumbers]{cvpr} 

\usepackage{booktabs}
\usepackage{graphicx}
\usepackage{multirow}
\usepackage[table]{xcolor}
\usepackage{tabularx}

\definecolor{cvprblue}{rgb}{0.21,0.49,0.74}
\usepackage[pagebackref,breaklinks,colorlinks,allcolors=cvprblue]{hyperref}

\def\confName{CVPR}
\def\confYear{2027}

\title{ViDM: Efficient and Order-Robust Parallel Visual Decision Making}

\author{%
Xudong Tan$^{1}$ \qquad
Peng Ye$^{2}$ \qquad
Chenyu Huang$^{1}$ \qquad
Yaoxin Yang$^{1}$ \qquad
Huafeng Qin$^{3}$ \qquad
Tao Chen$^{1,\dagger}$\\[4pt]
$^{1}$Fudan University \qquad
$^{2}$The Chinese University of Hong Kong\\
$^{3}$Chongqing Technology and Business University\\[2pt]
{\small $^{\dagger}$Corresponding author: \texttt{eetchen@fudan.edu.cn}}%
}

\begin{document}
\maketitle
\begin{abstract}
Visual decision-making demands accurate judgments
and timely responses.
Yet for multiple questions sharing the same visual input,
autoregressive multimodal large language models generate
answers token by token, accumulating sequential decoding
costs as question count grows and delaying completion
of the decision set.
Their predictions can also vary with question order.
We introduce ViDM, which transforms multi-question visual
inference into parallel decisions in a single forward pass,
improving both accuracy and inference efficiency.
ViDM organizes shared visual context, multiple questions,
and question-specific decision slots into one sequence,
reusing the native language-model head to produce
all answer distributions simultaneously.
A Decision-Query Mask aligns each decision query's direct
attention with shared context and its corresponding question,
while case-balanced multi-order training strengthens
robustness to question reordering. Across nine benchmarks spanning image, multi-image, and video tasks, ViDM improves Qwen3.5-4B's average accuracy from 73.49\% to 77.55\% while achieving $8.4\times$ the baseline's average end-to-end throughput. On GQA, where each evaluated image has 11.4 questions
on average, the end-to-end speedup reaches $20.3\times$.
ViDM also improves mean and worst-order accuracy
while reducing order-induced variation across multiple
benchmarks, with further validation on
LLaVA-OneVision-7B and InternVL3.5-4B.
\end{abstract}    
\section{Introduction}
\label{sec:intro}

\begin{figure}
    \centering
    \includegraphics[width=\linewidth]{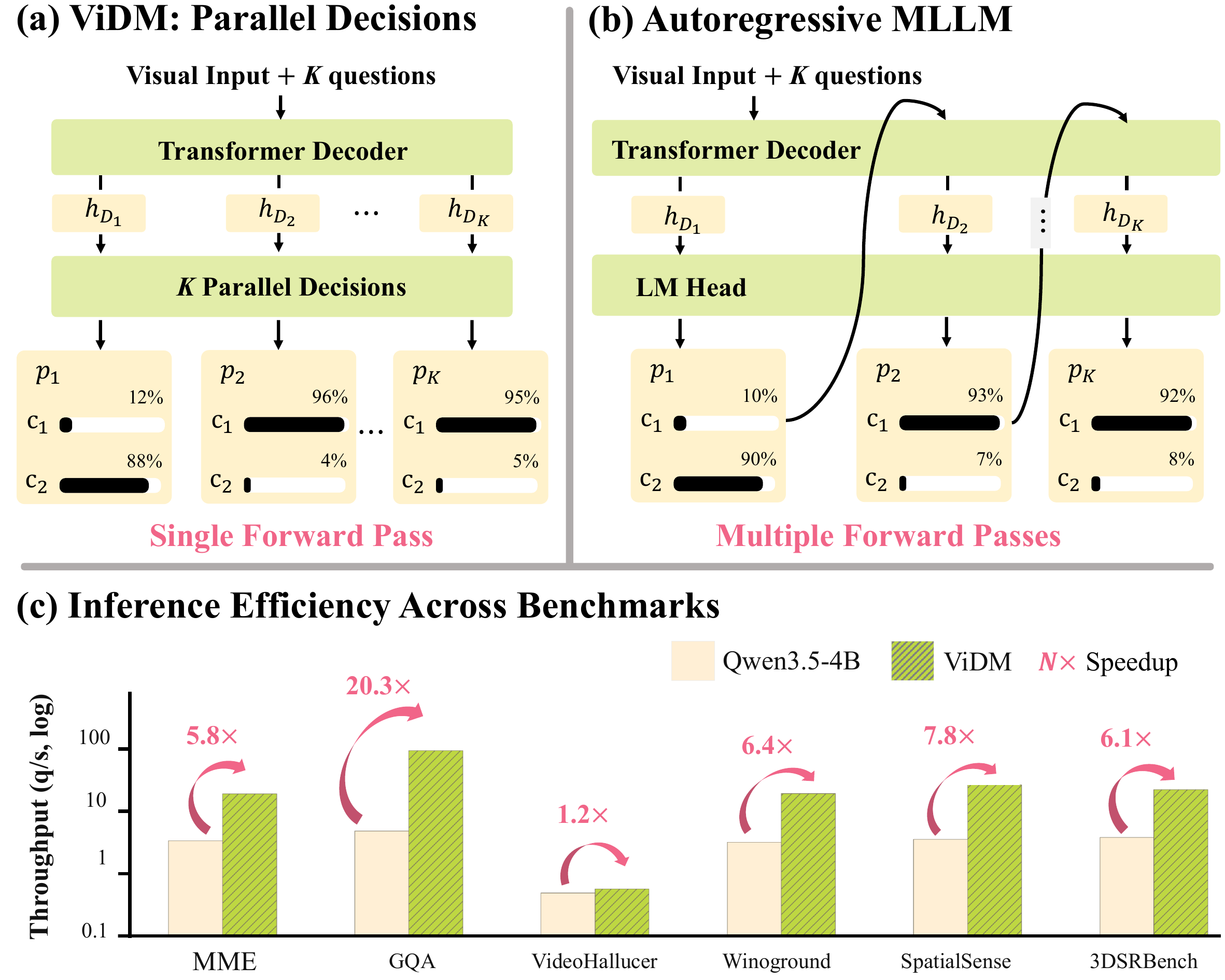}
    \caption{
    \textbf{ViDM: efficient parallel decisions for multiple questions.}
    (a) Given shared visual input and $K$ questions,
    ViDM predicts all answer distributions in a single forward pass.
    (b) Autoregressive MLLMs generate answers token by token
    through multiple forward passes.
    (c) ViDM achieves higher throughput than Qwen3.5-4B
    across benchmarks spanning image, multi-image, and video tasks
    (q/s, logarithmic scale).
    }
    \label{fig:ViDM_figure1}
\end{figure}

\begin{figure*}[ht]
    \centering
    \includegraphics[width=\textwidth]{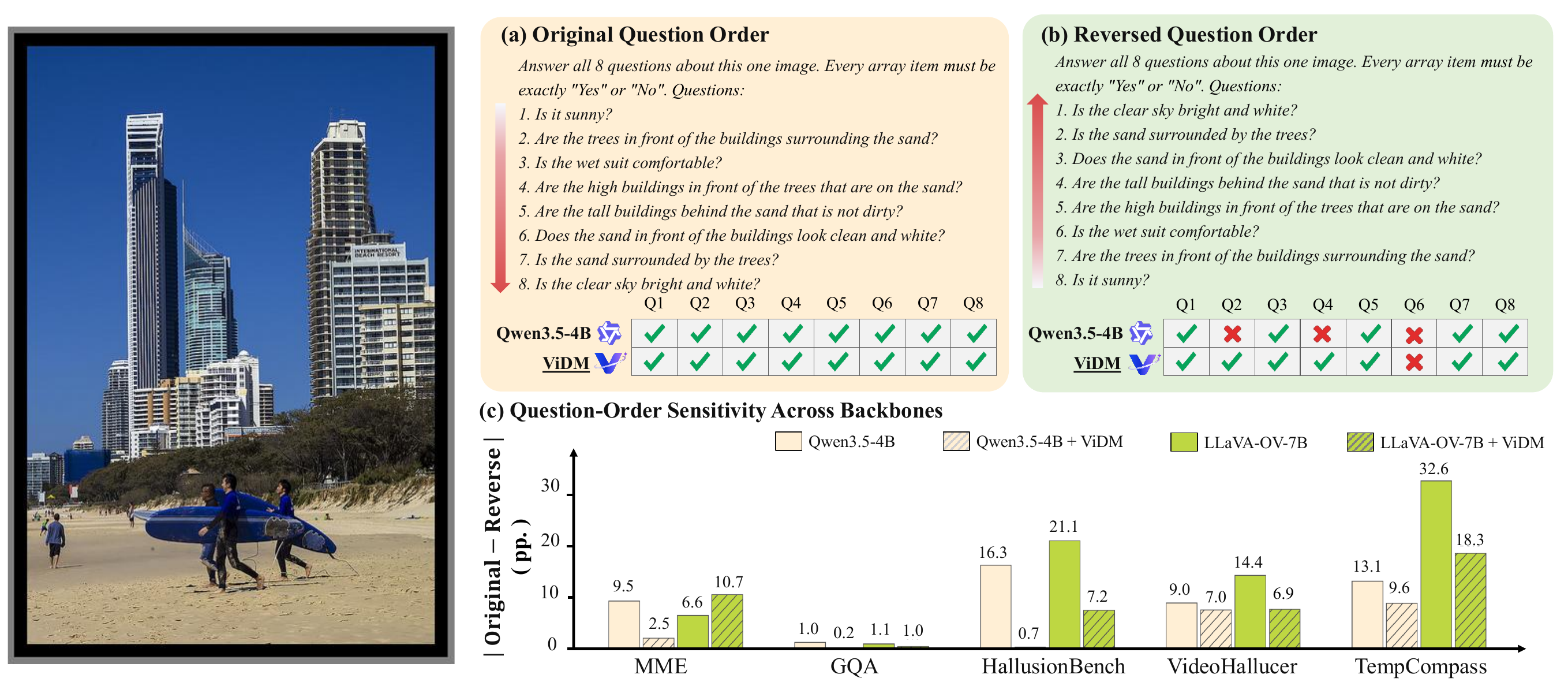}
    \caption{
    \textbf{Question-order sensitivity in multi-question visual reasoning.}
    (a--b) The same image and eight questions are presented
    in original and reversed orders.
    Despite unchanged visual evidence and question content,
    Qwen3.5-4B's answers vary with question order,
    while ViDM produces more consistent decisions.
    Green checks and red crosses denote correct and incorrect answers.
    (c) Absolute original--reverse accuracy gaps (pp.) show
    that this sensitivity is not an isolated case:
    it occurs across tasks in both Qwen3.5-4B and LLaVA-OneVision-7B.
    ViDM reduces these gaps in most comparisons.
    }
    \label{fig:ViDM_figure2}
\end{figure*}

Understanding a single image, image collection, or video
often requires multiple judgments, including verifying
visual content~\cite{guan2024hallusionbench}, identifying
spatial relations~\cite{ma20253dsrbench}, comparing
views~\cite{fu2024blink}, and understanding events over
time~\cite{fu2025video,li2024mvbench}.
Multimodal large language models (MLLMs) increasingly support
these perceptual and reasoning capabilities through
a unified language interface~\cite{li2024llava,bai2025qwen3,wang2025internvl3}.
These questions share visual evidence but have distinct
decision targets.
When a set of questions is available together,
the model should efficiently produce decisions
rather than construct a lengthy sequence of textual answers.

Autoregressive MLLMs organize these judgments as sequential
text generation.
As illustrated in Figure~\ref{fig:ViDM_figure1},
even when all questions share one request,
answers are generated token by token, accumulating
decoding cost with output length and question count.
Existing methods reduce computation through visual token
pruning and compression~\cite{chen2024image,yang2025visionzip,cao2024madtp},
spatiotemporal redundancy reduction~\cite{shen2026fastvid,shen2024longvu},
patch selection before visual encoding~\cite{shi2026attend},
and shared-prefix cache reuse~\cite{zheng2024sglang}.
Direct decision models further change how answers
are produced.
Visual Jev~\cite{yu2026visual} and Vev~\cite{vev2026}
reuse a shared prefix, batch isolated question branches,
and read candidate probabilities from the language-model head.
Valen~\cite{valen2026} combines shared-state computation with a Mixer
that fuses context, instruction, candidate, and
decision-position representations for scoring.
These methods demonstrate the efficiency potential of
non-generative decisions and make computation sharing
and readout organization central design choices
for multi-question inference.

Organizing multiple questions within one sequence
also determines their positions and contextual relationships.
Even with unchanged visual evidence and question content,
reordering questions can alter model predictions.
Figure~\ref{fig:ViDM_figure2} shows this sensitivity
across different MLLMs and visual tasks.
Related studies identify sensitivity to evidence
position~\cite{liu2024lost}, answer-option order~\cite{pezeshkpour2024large}, and multimodal
context arrangements~\cite{tan2024order}, and explore permutation-invariant
modeling of set-structured inputs~\cite{egressy2026set}.
For multi-question visual decisions, the challenge is
to share computation while aligning each readout
with its corresponding question and maintaining
robustness to reordering.

To this end, we introduce ViDM, which organizes
multi-question visual inference as parallel decisions
within a single forward pass.
ViDM packs shared visual context, multiple questions,
and their corresponding decision slots into one sequence.
A shared readout reuses the corresponding rows of
the backbone's native language-model head to produce
all answer distributions simultaneously.
To align each decision readout with its question,
we introduce a Decision-Query Mask
that restricts each decision query's direct attention
to shared context, its own question, and its own decision cue.
We further introduce Case-balanced Multi-order CE,
which supervises multiple permutations with
question-aligned labels and balances the training
contributions of individual cases.
This training encourages stable decisions as question
positions and surrounding context change.

We evaluate ViDM on nine benchmarks spanning image,
multi-image, and video tasks.
Compared with native Qwen3.5-4B, ViDM improves average
accuracy by 4.06 percentage points
and achieves $8.4\times$ average end-to-end throughput
on a single GPU.
Its throughput advantage grows with question count.
Across four question-reordering benchmarks,
ViDM improves both mean and worst-order accuracy
while reducing performance variation across orders.
The design also extends to LLaVA-OneVision-7B
and InternVL3.5-4B.
Furthermore, ViDM serves as a low-cost front-end
decision-maker that routes low-confidence questions
to Qwen3.5-27B, improving both accuracy and end-to-end
efficiency on GQA and HallusionBench. Our contributions are summarized as follows:
\begin{itemize}
    \item We identify question-order sensitivity across
    MLLMs and visual tasks: changing question order alters
    predictions despite unchanged visual evidence and
    question content. Together with sequential decoding
    overhead, this motivates efficient and order-robust
    multi-question decisions.

    \item We introduce ViDM for single-pass parallel decisions
    over shared visual input.
    Its Decision-Query Mask aligns each decision query's
    direct attention with its own question and shared context,
    while Case-balanced Multi-order CE applies question-specific
    supervision across different permutations to reduce
    question-order sensitivity.

    \item We evaluate ViDM across nine benchmarks spanning
    image, multi-image, and video tasks.
    On Qwen3.5-4B, ViDM improves average accuracy from
    73.49\% to 77.55\%, with reported average E2E throughput
    reaching $8.4\times$ that of the native baseline.
\end{itemize}
\begin{figure*}[!t]
    \centering
    \includegraphics[width=\textwidth]{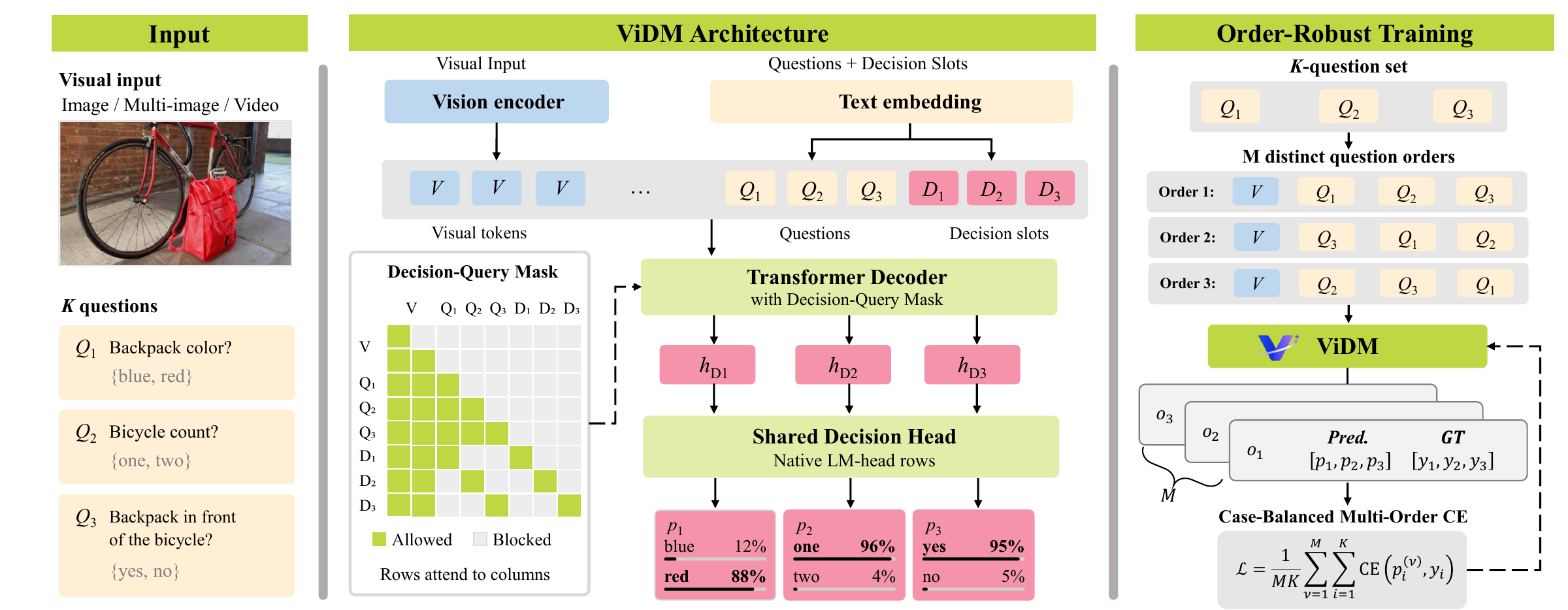}
    \caption{\textbf{Architecture and training pipeline of ViDM.}
    \textbf{(Left and middle)}
    Given an image, multiple images, or a video with $K$ questions,
    ViDM feeds shared visual tokens, question tokens, and $K$ decision
    slots into a Transformer decoder.
    A Decision-Query Mask restricts each decision query's direct
    attention to shared context, its own question, and its own cue.
    A shared decision head reuses native LM-head rows to map decision
    states to $K$ answer distributions in parallel, without
    autoregressive answer decoding.
    \textbf{(Right)}
    During training, the same question set is presented in $M$ distinct
    orders, with predictions aligned to their question-specific labels.
    Case-balanced Multi-order CE averages cross-entropy over questions
    and orders within each case, then weights cases equally,
    encouraging order-robust decisions.
    }
    \label{fig:vidm_framework}
\end{figure*}

\section{Related Work}
\label{sec:related-work}
\paragraph{Efficient Multimodal Inference.}
Existing methods improve multimodal inference efficiency
by reducing visual processing costs and accelerating
text decoding.
MADTP~\cite{cao2024madtp}, VisionZip~\cite{yang2025visionzip}, and AutoGaze~\cite{shi2026attend} reduce
input redundancy through alignment-guided pruning,
visual-token compression, and patch selection, respectively.
FastVLM~\cite{vasu2025fastvlm} lowers encoding latency and token count
through an efficient vision encoder, while LLaVA-Mini~\cite{zhang2025llava} combines modality pre-fusion with visual compression
to reduce the visual tokens processed by the language model.
However, reducing input-side costs leaves the sequential
dependencies of answer generation intact.
On the output side, Medusa~\cite{cai2024medusa} and EAGLE~\cite{li2024eagle} construct
candidate continuations through multiple prediction heads
or feature-level drafting, followed by target-model verification.
ViSpec~\cite{kang2026vispec} and DREAM~\cite{hu2026dream} further incorporate visual
information and target-model features into multimodal
speculative decoding.
These methods reduce sequential generation steps but
still construct answers through successive drafting
and verification rounds.
For multiple questions over shared visual input,
the overhead of progressively generating their answers remains.

\paragraph{Visual Decision Models.}
Visual scoring methods explore how to extract judgments
directly from multimodal representations.
Q-Align~\cite{wu2023q} and DeQA-Score~\cite{you2025teaching} assess visual quality
through rating-level or score-distribution prediction.
VQAScore~\cite{lin2024evaluating} measures image--text compatibility through
answer likelihoods, while LLaVA-Reward~\cite{zhou2025multimodal} derives
preference scores for generated images from hidden states.
LLaVA-Critic~\cite{xiong2025lllava} and Prometheus-Vision~\cite{lee2024prometheus} further
evaluate model responses through scores and textual feedback.
These formulations primarily assess visual content or
existing responses, rather than jointly answering
multiple questions over shared visual input.
For multi-question decisions, Visual Jev~\cite{yu2026visual} and Vev~\cite{vev2026} share a common prefix, execute isolated question branches
in batches, and read candidate probabilities from
the language-model head. Valen~\cite{valen2026} combines shared-state computation with a
Mixer head that fuses contextual and candidate-specific
representations.
However, jointly processing questions in a shared causal
sequence also couples decision representations to
question positions and surrounding context.
Efficient shared computation must therefore be paired
with question-aligned information access and supervision
across reordered inputs.
To this end, we propose ViDM, which reads multiple
decisions in parallel through the native language-model
head in one forward pass.
Its Decision-Query Mask constrains the direct attention
of decision queries, while case-balanced multi-order
training strengthens robustness to question reordering.

\section{Method}
\label{sec:Method}

ViDM formulates multi-question answering over shared visual
input as parallel decision-making in a single forward pass.
The central idea is to share visual computation while
constructing question-specific decision representations
that remain stable across question orders.
As illustrated in Fig.~\ref{fig:vidm_framework},
decision cues identify readout states, and a shared decision
head produces all answer distributions simultaneously.
A Decision-Query Mask controls each decision query's direct
attention, while case-balanced multi-order training learns
stable representations through supervision across permutations.

\subsection{Parallel Visual Decision-Making}
\label{sec:parallel_decisions}

Given visual input $v$ and $K$ questions
$\{q_i\}_{i=1}^{K}$, our goal is to predict their answers jointly.
The visual input can be an image, multiple images, or a video.
Native multi-question prompting shares visual context but
still represents answers as an autoregressively generated
token sequence, with later outputs conditioned on previously
generated answer tokens.
ViDM replaces this output process with joint readout from
multiple decision positions.

Using the backbone's native multimodal interface, we pack
the visual input, shared textual context $S$, questions,
and corresponding decision cues $D_i$:
\begin{equation}
    X = \operatorname{Pack}
    \left(v,S,\{q_i\}_{i=1}^{K},\{D_i\}_{i=1}^{K}\right).
    \label{eq:packed_input}
\end{equation}
Here, $\operatorname{Pack}$ constructs the input using the
backbone's prompt template.
All questions share the visual tokens, and each decision cue
provides a readout location for its corresponding question.
Let $d_i$ denote the readout position at the end of cue $D_i$.
The backbone produces all decision states in parallel as
\begin{equation}
    H=f_{\theta}(X;\mathbf M),
    \qquad
    h_i=H[d_i],
    \label{eq:decision_states}
\end{equation}
where $f_{\theta}$ is the multimodal backbone and $\mathbf M$
is the attention mask.
The states are computed together in one forward pass,
without feeding predicted answers back as generation inputs
for subsequent decisions.
For a candidate answer-label set $\mathcal A$, we extract
the corresponding weight rows and biases from the native
LM head to construct a shared decision head:
\begin{equation}
    \begin{aligned}
        p_i &=
        \operatorname{softmax}
        \left(W_{\mathcal A}h_i+b_{\mathcal A}\right),\\
        \hat y_i &= \arg\max_{a\in\mathcal A}p_i(a).
    \end{aligned}
    \label{eq:decision_readout}
\end{equation}
This readout reuses the pretrained answer mapping and
normalizes scores over candidate labels.
We freeze $W_{\mathcal A}$ and $b_{\mathcal A}$ during
training and update the backbone to form decision
representations suited to this mapping.
At inference, all $K$ answers are obtained directly
and in parallel from their states, sharing visual processing
and multimodal forward computation while eliminating
the sequential steps of token-by-token answer decoding.

\begin{figure}
    \centering
    \includegraphics[width=\linewidth]{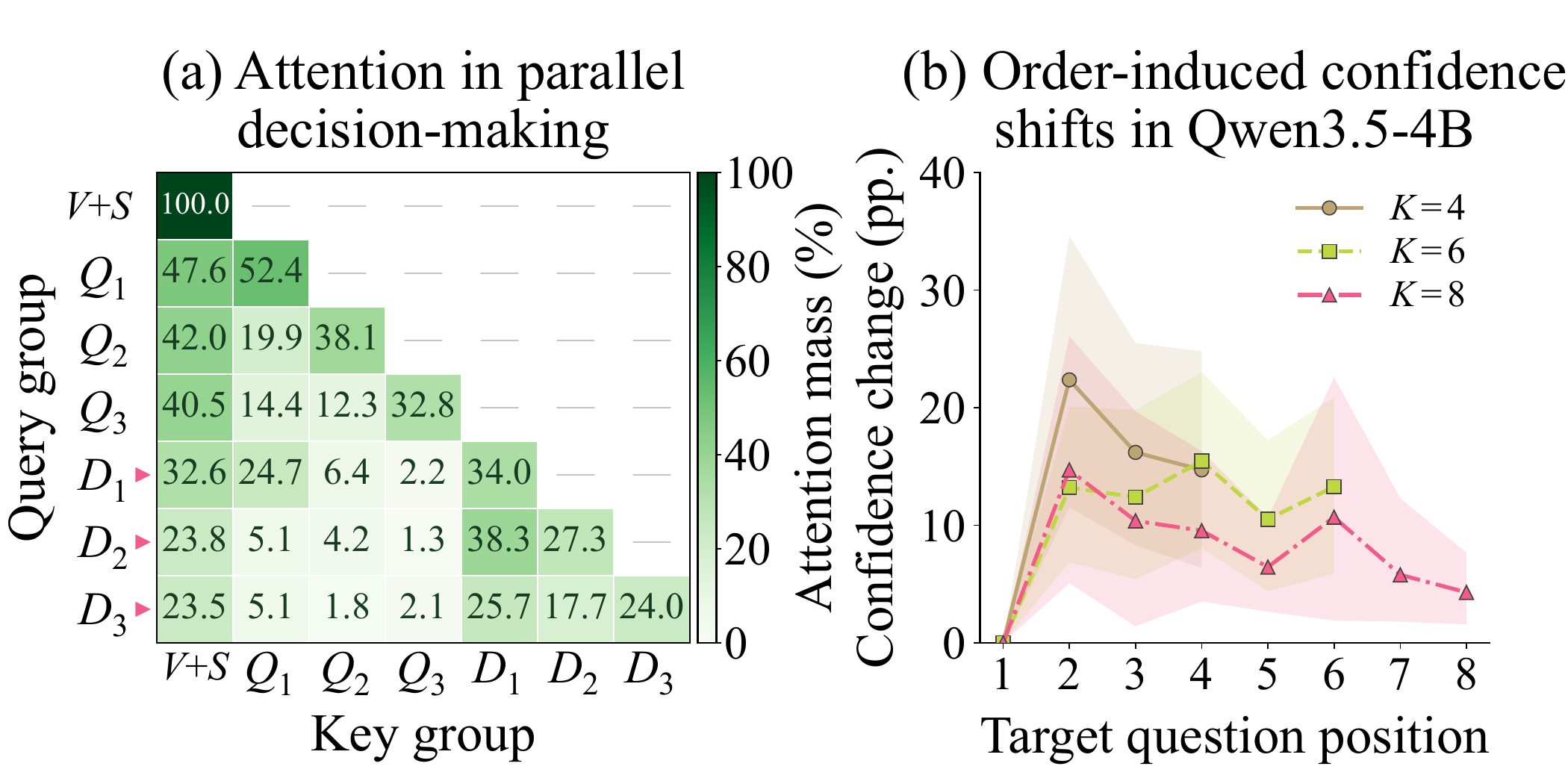}
    \caption{
    \textbf{Attention allocation and order sensitivity.}
    (a) At the decision layer (Layer~32) of the untrained,
    unmasked parallel model, decision-token regions
    (pink arrows) allocate substantial attention to other
    questions and earlier decision cues. $V{+}S$: visual/shared context; $D$: decision-cue spans.
    (b) Cyclically moving the same question to different
    positions changes answer confidence across $K$.
    Curves show mean absolute confidence changes from
    position~1, conditioned on preceding ground-truth answers.
    }
    \label{fig:ViDM_order_sensitivity}
\end{figure}

\subsection{Decision-Query Mask}
\label{sec:decision_mask}

Figure~\ref{fig:ViDM_order_sensitivity}(a) shows that,
before ViDM training and masking, the parallel model
allocates substantial attention to other questions and
earlier decision cues at the decision layer.
These cross-question and cross-cue paths allow decision
queries to directly access information belonging to
other questions.
Our Decision-Query Mask aligns each decision query's
direct attention with its corresponding question.

Let $\mathcal C$ contain the token indices of shared visual
and textual context, and let $\mathcal Q_i$ and $\mathcal D_i$
contain those of question $q_i$ and its decision cue.
For readout query $d_i$, the additive attention mask is
\begin{equation}
    \mathbf M_{d_i,t}=
    \begin{cases}
        0,
        & t\le d_i,\;
          t\in\mathcal C\cup\mathcal Q_i\cup\mathcal D_i,\\
        -\infty, & \text{otherwise}.
    \end{cases}
    \label{eq:decision_query_mask}
\end{equation}
The mask preserves causality and provides access to three
sources: shared visual and textual evidence, the question
defining the current task, and the cue identifying its
readout position.
Other questions and decision cues are excluded from
this query's direct attention.

We apply the constraint to readout-query rows in
full-attention layers; other query rows and linear-attention
layers retain the backbone's native computation.
Here, $D_i$ denotes an entire cue span, whereas $d_i$ is
the individual position from which the decision state
is extracted.
The structural constraint therefore acts on the queries
used for answer readout while preserving representation
computation for shared context and question tokens.

\definecolor{gainpink}{HTML}{F16588}

\begin{table*}[t]
\centering
\small
\renewcommand{\arraystretch}{1.25}
\setlength{\tabcolsep}{3.5pt}
\caption{
Performance (\%) and single-H200 efficiency on nine benchmarks.
Performance averages Original/Reverse accuracy.
Avg. Perf. and Avg. E2E report equal-dataset averages of
performance and end-to-end throughput, respectively.
Bold and underlining indicate the best and second-best values;
Qwen3.5-27B is excluded only from performance ranking.
\newline
\textbf{Image Task:}
IT-1: GQA~\cite{hudson2019gqa};
IT-2: HallusionBench~\cite{guan2024hallusionbench};
IT-3: MME~\cite{fu2026mme};
IT-4: SpatialSense~\cite{yang2019spatialsense};
IT-5: VSR~\cite{liu2023visual};
IT-6: 3DSRBench~\cite{ma20253dsrbench}.
\newline
\textbf{Multi-image Task:} MIT-1: Winoground~\cite{thrush2022winoground}.
\quad
\textbf{Video Task:} VT-1: VideoHallucer~\cite{wang2024videohallucer}; VT-2: TempCompass~\cite{liu2024tempcompass}.
}
\label{tab:main-nine}
\resizebox{\textwidth}{!}{%
\begin{tabular}{l*{11}{c}}
\toprule
\multirow{2}{*}[-2pt]{Method}
& \multicolumn{6}{c}{Image Task}
& \multicolumn{1}{c}{Multi-image Task}
& \multicolumn{2}{c}{Video Task}
& \multicolumn{2}{c}{Overall} \\
\cmidrule(lr){2-7}
\cmidrule(lr){8-8}
\cmidrule(lr){9-10}
\cmidrule(lr){11-12}
& IT-1 & IT-2 & IT-3 & IT-4 & IT-5 & IT-6
& MIT-1 & VT-1 & VT-2
& Avg. Perf.$\uparrow$ & Avg. E2E (q/s)$\uparrow$ \\
\midrule
Qwen3.5-27B
& 81.14 & 91.21 & 92.21 & 71.35 & 88.47 & 60.08
& 86.53 & 86.41 & 83.86 & 82.36 & 2.10 \\
Qwen3.5-9B
& 70.63 & 81.64 & 73.44 & 67.40
& \underline{85.68} & \underline{56.75}
& 76.12 & \textbf{83.43} & \textbf{78.71}
& 74.87 & 3.01 \\
Qwen3.5-4B
& 65.61 & 80.03 & 85.87 & 64.21 & 82.54
& \textbf{56.90} & 70.91 & 78.46 & 76.92
& 73.49 & 3.56\,{\scriptsize($1.0\times$)} \\
InternVL3.5-4B
& 76.34 & 76.26 & 85.07 & 60.82 & 75.69 & 53.65
& 68.06 & 78.84 & 73.11 & 71.98 & 5.35 \\
LLaVA-OneVision-7B
& 61.90 & 72.73 & 76.47 & 60.18 & 73.21 & 52.94
& 56.72 & 70.68 & 65.13 & 65.55 & 7.13 \\
Visual-Jev (4B)
& \underline{80.45} & 78.28 & 87.62 & 60.33 & 80.14 & 52.06
& 67.94 & 79.32 & 71.40 & 73.06 & 16.64 \\
Valen-4B
& \textbf{81.49} & \underline{85.95} & 88.42
& \underline{68.12} & 83.94 & 52.62
& \underline{77.34} & 80.68 & 77.40
& \underline{77.33} & \underline{27.73} \\
Vev-4B
& 80.25 & 83.36 & \underline{88.63}
& \textbf{69.67} & \textbf{86.02} & 55.56
& \textbf{77.62} & 80.54 & 73.93
& 77.29 & 10.32 \\
\midrule
\rowcolor{gray!12}
Qwen3.5-4B + ViDM
& 79.73 & \textbf{86.45} & \textbf{93.26}
& 66.20 & 84.04 & 54.05
& 75.59 & \underline{81.08} & \underline{77.54}
& \textbf{77.55}\,{\scriptsize\textcolor{gainpink}{(+4.06)}}
& \textbf{29.95}\,{\scriptsize\textcolor{gainpink}{($8.4\times$)}} \\
\bottomrule
\end{tabular}%
}
\end{table*}

\subsection{Case-balanced Multi-order Training}
\label{sec:multi_order_training}

Figure~\ref{fig:ViDM_order_sensitivity}(b) shows
order-dependent changes in answer confidence even when
preceding answers are supplied by ground-truth labels.
Thus, order sensitivity persists without preceding
prediction errors.
Since reordering changes question positions and
context arrangements, we train the model to recover
question-specific answers across these variations.

For case $c$ with $K_c$ questions, we construct $M_c$
distinct permutations.
Each permutation preserves the visual input, question
content, and labels, while pairing decision positions
with questions in the current order.
Let $\pi_c^{(m)}(j)$ identify the original question occupying
position $j$ in permutation $m$, and let $p_{c,j}^{(m)}$
be the prediction at that position.
The loss for each permutation is
\begin{equation}
    \ell_c^{(m)}
    =
    -\frac{1}{K_c}
    \sum_{j=1}^{K_c}
    \log p_{c,j}^{(m)}
    \left(y_{c,\pi_c^{(m)}(j)}\right).
    \label{eq:order_loss}
\end{equation}
This objective assigns supervision by question identity
rather than fixed position and averages losses over
questions within the permutation.
We then average over permutations within each case and
weight the $B$ cases in a batch equally:
\begin{equation}
    \mathcal L
    =
    \frac{1}{B}
    \sum_{c=1}^{B}
    \frac{1}{M_c}
    \sum_{m=1}^{M_c}
    \ell_c^{(m)}.
    \label{eq:case_balanced_multi_order}
\end{equation}
The two normalization factors account for question count
and permutation count, preventing cases with more questions
or orders from receiving greater total training weight.
Each case consequently contributes as one multi-question
learning unit grounded in shared visual input.

The Decision-Query Mask specifies direct information
access, while multi-order supervision learns to make
the same question-specific decision across positions
and context arrangements.
These components complement each other through structure
and supervision.
Multiple orders are used only during training:
inference produces all answers under the supplied order,
without permutation ensembling.

\definecolor{gainpink}{HTML}{F16588}

\begin{figure*}[t]
    \centering
    \includegraphics[width=\linewidth]{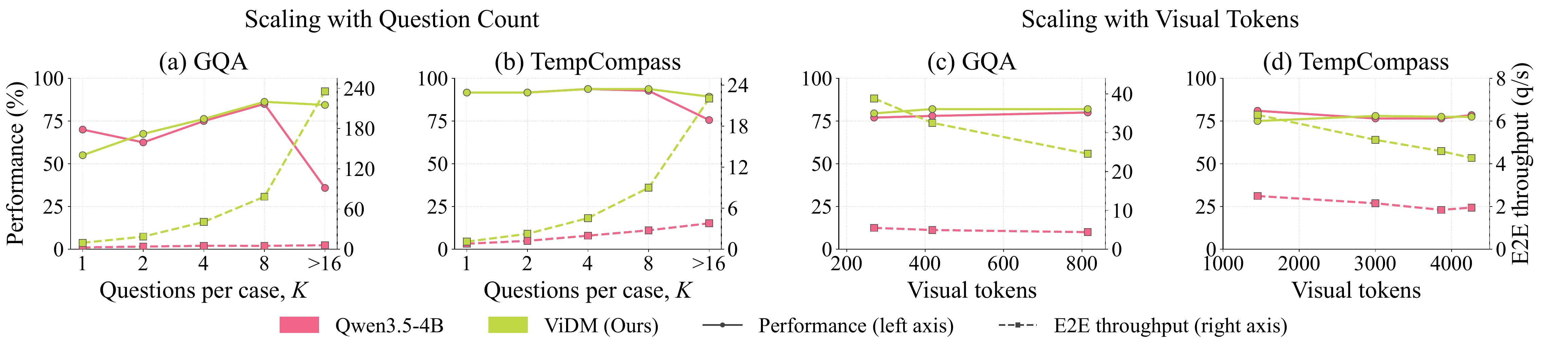}
    \caption{
    \textbf{Efficiency and scalability.}
    Performance and end-to-end throughput of ViDM and autoregressive
    Qwen3.5-4B on GQA and TempCompass.
    (a--b) ViDM's throughput advantage widens as the question count $K$ increases.
    (c--d) With $K=4$, ViDM maintains higher throughput across visual-token
    budgets, demonstrating the persistent efficiency benefits of parallel decisions.
    }
    \label{fig:ViDM_efficiency_scalability}
\end{figure*}

\section{Experiments}
\label{sec:experiments}

We evaluate ViDM in terms of prediction performance,
efficiency and scalability, order robustness, decision reliability,
and transfer across backbones.

\subsection{Experimental Setup}
\label{sec:experimental_setup}

\paragraph{Benchmarks and baselines.}
We evaluate ViDM on nine benchmarks spanning image, multi-image,
and video tasks, as listed in Table~\ref{tab:main-nine}.
We adopt a unified multi-question decision protocol, grouping
questions sharing visual input into a case.
Baselines include Qwen3.5-4B/9B/27B~\cite{qwen3.5}, InternVL3.5-4B~\cite{wang2025internvl3},
LLaVA-OneVision-7B~\cite{li2024llava}, Visual-Jev~\cite{yu2026visual}, Valen~\cite{valen2026}, and Vev~\cite{vev2026}.
Unless otherwise specified, ViDM uses Qwen3.5-4B as its backbone.

\paragraph{Implementation details.}
We train ViDM on a mixture of 57,857 image and video cases,
with up to 16 questions per training case.
We construct up to 10 distinct question orders per case,
depending on its question count, and optimize
Case-balanced Multi-order CE.
Training uses two H200 GPUs, BF16, a global batch size of 16,
and a 15-epoch schedule.
The learning rates are $10^{-6}$ for the language module
and $2\times10^{-7}$ for the vision module.
We use AdamW with 5\% warmup and cosine learning-rate decay.

\paragraph{Evaluation metrics.}

We compute accuracy within each case and average cases equally.
Overall performance and throughput are equal-weight averages
across datasets.
Efficiency is measured on one H200 with BF16 and batch size 1.
End-to-end time includes media decoding, preprocessing, transfer,
model execution, and answer parsing, excluding loading and warmup.
Throughput is the number of answered questions divided
by end-to-end time.

\subsection{Main Results}
\label{sec:main_results}

Table~\ref{tab:main-nine} shows that ViDM improves both accuracy
and throughput for multi-question decisions.
Compared with native Qwen3.5-4B, ViDM raises average accuracy
from 73.49\% to 77.55\%, a gain of 4.06 percentage points,
while increasing average end-to-end throughput from
3.56 to 29.95 questions/s, or $8.4\times$ the baseline.
Accuracy improves on eight of nine benchmarks, including
gains of 14.12 and 7.39 points on GQA and MME, respectively.

Among the compared 4B--9B models, ViDM achieves the highest
average accuracy and throughput.
It also exceeds Valen and Vev in average accuracy while
processing more questions per second.
These results demonstrate that replacing token-by-token
answer generation with parallel decisions over shared visual
context improves both prediction quality and processing capacity.

\subsection{Efficiency and Scalability}
\label{sec:efficiency_scalability}

\paragraph{Scaling with question count.}
We vary $K$ over $1,2,4,8,$ and $>16$ on fixed visual cases
from GQA and TempCompass.
Figure~\ref{fig:ViDM_efficiency_scalability}(a--b) shows
that ViDM's throughput advantage grows with $K$:
end-to-end speedup increases from $3.7\times$ to $39.5\times$
on GQA and from $1.4\times$ to $5.8\times$ on TempCompass.
Native generation latency grows with the number of questions,
whereas ViDM maintains relatively stable model latency
over the tested range by producing multiple decisions
in one forward pass.
At $K>16$, ViDM achieves 84.41\% and 89.28\% accuracy,
respectively, demonstrating effective scaling to larger
question sets.

\paragraph{Scaling with visual tokens.}
We fix $K=4$ and vary spatial pixel budgets on
50 randomly sampled cases per dataset,
plotting the actual visual-token counts.
Figure~\ref{fig:ViDM_efficiency_scalability}(c--d) shows
higher ViDM throughput at every displayed budget,
with speedups of $5.6$--$7.1\times$ on GQA and
$2.2$--$2.5\times$ on TempCompass.
As visual inputs grow, increasing visual-processing cost
narrows the relative benefit of parallel decisions,
but the throughput advantage persists.
Meanwhile, GQA accuracy increases from 79.5\% to 82.0\%,
and TempCompass accuracy remains within 75.0\%--78.0\%.

\subsection{Order Robustness and Decision Reliability}
\label{sec:robustness_reliability}

\begin{table}[t]
\centering
\small
\renewcommand{\arraystretch}{1.25}
\setlength{\tabcolsep}{2pt}
\caption{
\textbf{Performance and stability under question reordering.}
O/R/C/S denote original, reversed, cyclic, and shuffled orders.
Avg. averages accuracy (\%) across orders;
Worst and Range average the per-case minimum accuracy
and accuracy range (pp.), respectively.
Bold marks better results within each dataset.
}
\label{tab:order-robustness}
\begin{tabular*}{\columnwidth}{@{\extracolsep{\fill}}l*{7}{c}@{}}
\toprule
Method & O$\uparrow$ & R$\uparrow$
& C$\uparrow$ & S$\uparrow$
& Avg.$\uparrow$ & Worst$\uparrow$ & Range$\downarrow$ \\
\midrule
\rowcolor{gray!12}
\multicolumn{8}{l}{\textbf{GQA}} \\
Qwen3.5-4B
& 65.0 & 66.1 & 65.1 & 65.5
& 65.4 & 54.5 & 21.2 \\
+ ViDM
& \textbf{79.8} & \textbf{79.6}
& \textbf{80.2} & \textbf{79.4}
& \textbf{79.8} & \textbf{75.9} & \textbf{7.3} \\
\midrule
\rowcolor{gray!12}
\multicolumn{8}{l}{\textbf{HallusionBench}} \\
Qwen3.5-4B
& \textbf{88.2} & 71.9 & 73.4 & 82.3
& 78.9 & 67.6 & 24.6 \\
+ ViDM
& 86.1 & \textbf{86.8}
& \textbf{86.3} & \textbf{84.9}
& \textbf{86.0} & \textbf{77.4} & \textbf{16.7} \\
\midrule
\rowcolor{gray!12}
\multicolumn{8}{l}{\textbf{SpatialSense}} \\
Qwen3.5-4B
& 63.0 & 65.4 & 64.8 & 65.6
& 64.7 & 50.0 & 28.4 \\
+ ViDM
& \textbf{65.8} & \textbf{66.6}
& \textbf{66.4} & \textbf{66.6}
& \textbf{66.4} & \textbf{57.9} & \textbf{16.2} \\
\midrule
\rowcolor{gray!12}
\multicolumn{8}{l}{\textbf{VideoHallucer}} \\
Qwen3.5-4B
& 83.0 & 73.8 & 73.9 & 79.0
& 77.4 & 71.0 & 13.8 \\
+ ViDM
& \textbf{84.6} & \textbf{77.6}
& \textbf{78.9} & \textbf{81.1}
& \textbf{80.5} & \textbf{75.5} & \textbf{11.1} \\
\bottomrule
\end{tabular*}
\end{table}

\begin{figure}
    \centering
    \includegraphics[width=\linewidth]{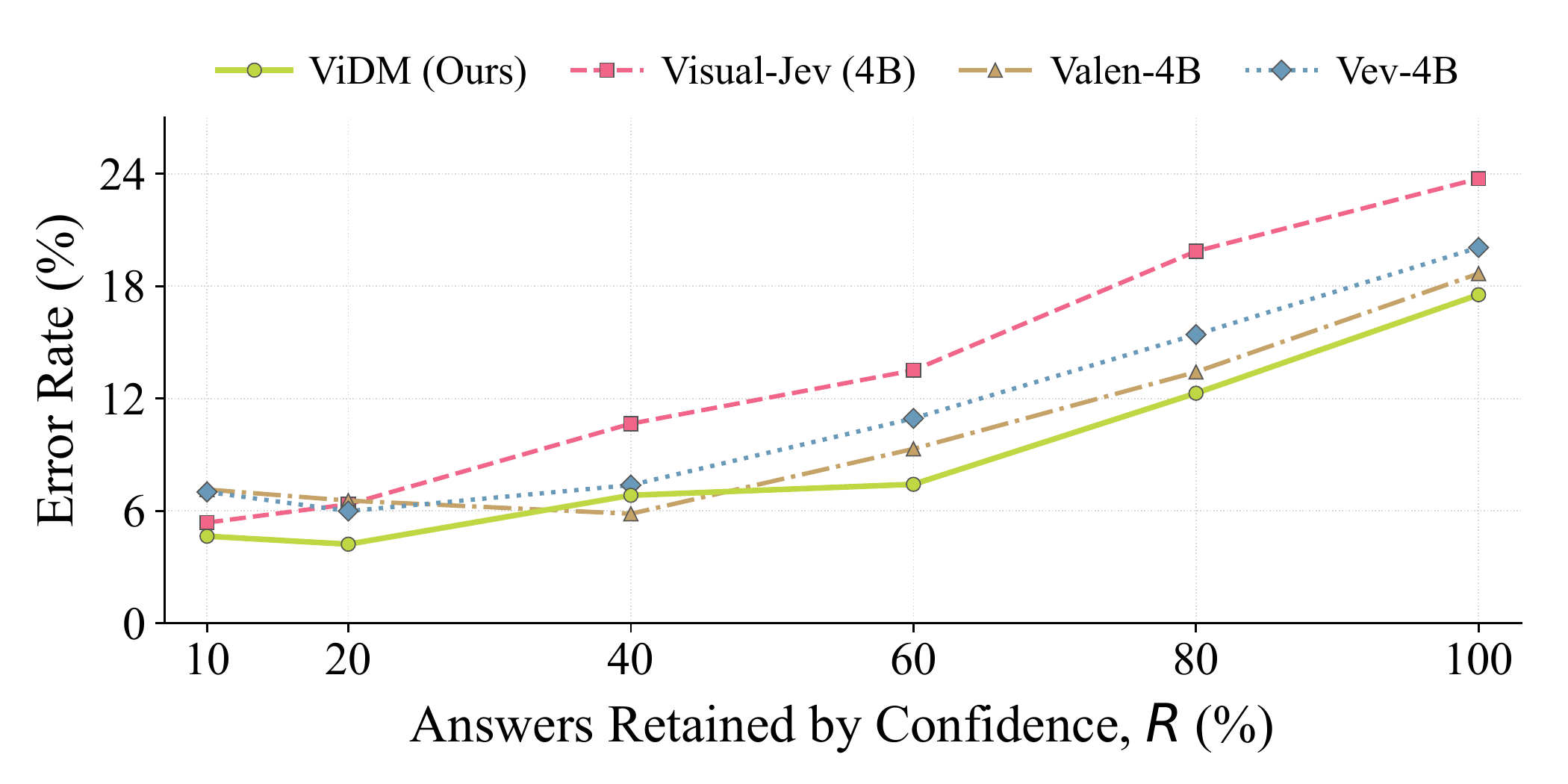}
    \caption{
    \textbf{Confidence-based decision reliability.}
    Models answer all questions in cases across multiple datasets.
    $R$ (\%) denotes the fraction of answers retained in descending
    confidence order within each dataset.
    Error rates among retained answers are averaged equally across
    datasets. At the same $R$, lower error rates indicate more
    reliable high-confidence answers.
    }
    \label{fig:ViDM_reliability}
\end{figure}

\paragraph{Robustness to question reordering.}
We evaluate original, reversed, cyclic, and shuffled orders
on four benchmarks.
Table~\ref{tab:order-robustness} shows that ViDM improves
mean and case-averaged worst-order accuracy while reducing
the case-averaged accuracy range on all four datasets.
On GQA, worst-order accuracy increases from 54.5\% to 75.9\%,
and the range decreases from 21.2 to 7.3 percentage points,
demonstrating reduced order sensitivity relative to
native Qwen3.5-4B.

\paragraph{Confidence-based decision reliability.}
We analyze decision reliability on GQA, HallusionBench,
MME, SpatialSense, and VideoHallucer. We compute error among each model's top-$R\%$ confident
answers within each dataset and average datasets equally.
ViDM achieves the lowest error at five of six retention levels
(Figure~\ref{fig:ViDM_reliability}).
Its error decreases from 17.54\% at full coverage to 4.22\%
at 20\% retention, showing a reliability advantage over
the three parallel decision models in these tasks.

\paragraph{ViDM-guided Qwen3.5-27B inference.}
ViDM acts as a low-cost front-end decision-maker
for multi-question visual tasks. R50 and R75 use nominal answer-retention rates of
50\% and 75\%, respectively, accepting ViDM's
highest-confidence answers and routing the remaining
questions to Qwen3.5-27B.
This reduces the answer-decoding workload of the larger model.
In Table~\ref{tab:vidm-guided-27b}, these settings achieve
$1.72$--$2.86\times$ E2E speedups on GQA and HallusionBench
while improving accuracy by $0.25$--$1.68$ percentage points.
R90 further increases throughput by accepting more ViDM answers.
On VideoHallucer, R50 and R75 improve accuracy by
$0.33$ and $1.00$ points, respectively, but do not accelerate
E2E inference; the substantial visual-token workload can
offset the savings in answer decoding.

\newcommand{\cascadeEtoE}[2]{%
    \mbox{#1\,{\scriptsize\textcolor{gainpink}{($#2\times$)}}}%
}

\begin{table}[t]
\centering
\small
\renewcommand{\arraystretch}{1.25}
\setlength{\tabcolsep}{1.5pt}
\caption{
\textbf{ViDM-guided inference with Qwen3.5-27B.}
ViDM (4B) serves as a low-cost front-end decision-maker.
R50/R75/R90 use ViDM to answer a target 50/75/90\%
of questions and route the rest to Qwen3.5-27B.
Task identifiers follow Table~\ref{tab:main-nine}.
We report original-order accuracy (\%) and E2E throughput (q/s).
Bold marks the best accuracy.
}
\label{tab:vidm-guided-27b}
\begin{tabularx}{\columnwidth}{
    l*{3}{c>{\centering\arraybackslash}X}
}
\toprule
\multirow{2}{*}[-1pt]{Method}
& \multicolumn{2}{c}{IT-1}
& \multicolumn{2}{c}{IT-2}
& \multicolumn{2}{c}{VT-1} \\
\cmidrule(lr){2-3}
\cmidrule(lr){4-5}
\cmidrule(lr){6-7}
& Acc.$\uparrow$ & E2E$\uparrow$
& Acc.$\uparrow$ & E2E$\uparrow$
& Acc.$\uparrow$ & E2E$\uparrow$ \\
\midrule
\rowcolor{gray!12}
ViDM
& 78.4 & 129.5
& 88.4 & 23.6
& 82.3 & 0.6 \\
Q.-3.5-27B
& 79.5 & \cascadeEtoE{3.4}{1.0}
& 91.3 & \cascadeEtoE{1.9}{1.0}
& 84.5 & \cascadeEtoE{0.4}{1.0} \\
\midrule
R50
& 79.8 & \cascadeEtoE{6.0}{1.8}
& \textbf{92.6} & \cascadeEtoE{3.2}{1.7}
& 84.8 & \cascadeEtoE{0.3}{0.8} \\
R75
& \textbf{81.1} & \cascadeEtoE{8.7}{2.6}
& 91.5 & \cascadeEtoE{5.3}{2.9}
& \textbf{85.5} & \cascadeEtoE{0.3}{0.8} \\
R90
& 79.4 & \cascadeEtoE{16.7}{5.0}
& 89.6 & \cascadeEtoE{8.0}{4.3}
& 83.0 & \cascadeEtoE{0.4}{1.0} \\
\bottomrule
\end{tabularx}
\end{table}

\subsection{Backbone and Modality Generalization}
\label{sec:backbone_generalization}

\begin{table}[t]
\centering
\small
\renewcommand{\arraystretch}{1.25}
\setlength{\tabcolsep}{2pt}
\caption{
\textbf{ViDM across backbones on image and video tasks.}
Task identifiers follow Table~\ref{tab:main-nine}.
Performance (\%) averages Original/Reverse accuracy.
Avg. and E2E report equal-dataset averages over the four
benchmarks of performance and single-H200 end-to-end
throughput (q/s), respectively.
Bold marks better results within each backbone.
}
\label{tab:backbone-modality}
\begin{tabular*}{\columnwidth}{@{\extracolsep{\fill}}l*{6}{c}@{}}
\toprule
Method
& IT-1$\uparrow$
& IT-2$\uparrow$
& IT-4$\uparrow$
& VT-1$\uparrow$
& Avg.$\uparrow$
& E2E$\uparrow$ \\
\midrule
\rowcolor{gray!12}
\multicolumn{7}{l}{\textbf{LLaVA-OneVision-7B}} \\
Native
& 61.90 & 72.73 & \textbf{60.18} & 70.68
& 66.37
& 8.52\,{\scriptsize($1.0\times$)} \\
+ ViDM
& \textbf{81.14} & \textbf{76.24}
& 60.08 & \textbf{74.51}
& \textbf{72.99}
& \textbf{22.93}\,{\scriptsize\textcolor{gainpink}{($2.7\times$)}} \\
\midrule
\rowcolor{gray!12}
\multicolumn{7}{l}{\textbf{InternVL3.5-4B}} \\
Native
& \textbf{76.34} & 76.26 & 60.82 & 78.84
& 73.07
& 5.76\,{\scriptsize($1.0\times$)} \\
+ ViDM
& 75.05 & \textbf{79.09}
& \textbf{61.24} & \textbf{79.47}
& \textbf{73.71}
& \textbf{16.96}\,{\scriptsize\textcolor{gainpink}{($2.9\times$)}} \\
\bottomrule
\end{tabular*}
\end{table}

We apply ViDM to LLaVA-OneVision-7B and InternVL3.5-4B
and evaluate GQA, HallusionBench, SpatialSense,
and VideoHallucer.
As shown in Table~\ref{tab:backbone-modality},
ViDM improves LLaVA-OneVision-7B's average accuracy
from 66.37\% to 72.99\% and increases average end-to-end
throughput from 8.52 to 22.93 questions/s ($2.7\times$).
For InternVL3.5-4B, average accuracy improves
from 73.07\% to 73.71\%, while throughput increases
from 5.76 to 16.96 questions/s ($2.9\times$).
Both backbones improve on HallusionBench and VideoHallucer,
showing that ViDM's parallel decision design transfers
across model architectures and image/video inputs.

\subsection{Ablation Studies}
\label{sec:ablation}

\begin{table*}[t]
\centering
\small
\renewcommand{\arraystretch}{1.1}
\setlength{\tabcolsep}{3.5pt}
\caption{
\textbf{Ablation of the Decision-Query Mask and multi-order training.}
We report mean Original/Reverse accuracy (\%) on nine benchmarks.
Task identifiers follow Table~\ref{tab:main-nine}.
MO denotes multi-order training; Avg. averages datasets equally.
The four ablation configurations use the same training data
and number of epochs.
Bold marks the best result in each column.
}
\label{tab:ablation}
\resizebox{\textwidth}{!}{%
\begin{tabular}{l*{12}{c}}
\toprule
\multirow{2}{*}[-2pt]{Method}
& \multirow{2}{*}[-2pt]{Mask}
& \multirow{2}{*}[-2pt]{MO}
& \multicolumn{6}{c}{Image Task}
& \multicolumn{1}{c}{Multi-image Task}
& \multicolumn{2}{c}{Video Task}
& \multirow{2}{*}[-2pt]{Avg.$\uparrow$} \\
\cmidrule(lr){4-9}
\cmidrule(lr){10-10}
\cmidrule(lr){11-12}
& & &
IT-1 & IT-2 & IT-3 & IT-4 & IT-5 & IT-6
& MIT-1 & VT-1 & VT-2 & \\
\midrule
Qwen3.5-4B (baseline)
& --- & ---
& 65.61 & 80.03 & 85.87 & 64.21
& \textbf{82.54} & \textbf{56.90}
& 70.91 & 78.46 & \textbf{76.92} & 73.49 \\
\midrule
Parallel baseline
& $\times$ & $\times$
& 78.62 & 78.52 & 82.83 & 65.00 & 82.08 & 56.19
& 68.94 & 78.14 & 74.52 & 73.87 \\
Mask only
& $\checkmark$ & $\times$
& \textbf{78.82} & 80.15 & 84.10 & 65.94 & 81.84 & 55.79
& 69.34 & 79.87 & 76.15 & 74.67 \\
Multi-order only
& $\times$ & $\checkmark$
& 78.53 & 84.61 & \textbf{92.99} & 66.44 & 82.35 & 56.83
& 70.09 & 80.25 & 75.74 & 76.42 \\
\midrule
\rowcolor{gray!12}
ViDM
& $\checkmark$ & $\checkmark$
& 78.56 & \textbf{85.18} & 92.94
& \textbf{66.71} & 82.46 & 56.43
& \textbf{71.50} & \textbf{80.76} & 76.87
& \textbf{76.82} \\
\bottomrule
\end{tabular}%
}
\end{table*}

\begin{table}[t]
\centering
\small
\renewcommand{\arraystretch}{1.25}
\setlength{\tabcolsep}{2pt}
\caption{
\textbf{Ablation of question-order robustness.}
Results are averaged equally over IT-1, IT-2, IT-4, and VT-1.
Across four question orders, Avg., Worst, and Range report
case-averaged mean accuracy (\%), minimum accuracy (\%),
and accuracy range (pp.), respectively.
MO denotes multi-order training.
Bold marks the best results across all evaluated configurations.
}
\label{tab:ablation-robustness}
\begin{tabularx}{\columnwidth}{
    l*{5}{>{\centering\arraybackslash}X}
}
\toprule
Method
& Mask
& MO
& Avg.$\uparrow$
& Worst$\uparrow$
& Range$\downarrow$ \\
\midrule
Qwen3.5-4B
& --- & ---
& 71.61 & 60.78 & 21.97 \\
\midrule
Parallel baseline
& $\times$ & $\times$
& 74.08 & 64.60 & 20.32 \\
Mask only
& $\checkmark$ & $\times$
& 75.33 & 66.32 & 18.98 \\
Multi-order only
& $\times$ & $\checkmark$
& 77.16 & 69.29 & 15.87 \\
\midrule
\rowcolor{gray!12}
ViDM
& $\checkmark$ & $\checkmark$
& \textbf{77.43} & \textbf{70.06} & \textbf{14.68} \\
\bottomrule
\end{tabularx}
\end{table}

\paragraph{Experimental setup.}
We evaluate four configurations on Qwen3.5-4B:
a parallel baseline, Mask only, multi-order training only,
and full ViDM.
The parallel baseline retains ViDM's shared visual context,
question-specific decision slots, and readout from frozen
native LM-head rows.
It predicts all answers in one forward pass using native
causal attention and is trained with a fixed question order.
All four configurations are initialized from the same
pretrained checkpoint and trained on the same 5K cases
for 10 epochs with the case-balanced objective.
Fixed-order configurations repeat the canonical order
to match the number of training views.
We evaluate the final checkpoints.
For nine-benchmark accuracy, we use the original
autoregressive Qwen3.5-4B results from
Table~\ref{tab:main-nine} as the reference.

\paragraph{Performance contributions.}
Table~\ref{tab:ablation} shows the contributions of both
components to average accuracy across nine benchmarks.
Adding the Decision-Query Mask to the parallel baseline
raises accuracy from 73.87\% to 74.67\%.
Multi-order training provides the larger gain, reaching
76.42\%, an improvement of 2.55 percentage points
over the parallel baseline.
Combining both components yields 76.82\%,
exceeding multi-order training alone by 0.40 points
and the parallel baseline by 2.95 points.
Full ViDM also surpasses the native baseline (73.49\%)
by 3.33 percentage points.

\paragraph{Question-order robustness.}
Table~\ref{tab:ablation-robustness} evaluates stability
across four question orders.
Multi-order training raises case-averaged worst-order
accuracy from 64.60\% to 69.29\% and reduces the
accuracy range from 20.32 to 15.87 percentage points.
Adding the Mask further increases worst-order accuracy
to 70.06\% and narrows the range to 14.68 points,
corresponding to an additional improvement of 0.77 points
and a reduction of 1.19 points, respectively.
Thus, multi-order supervision drives substantial gains
in accuracy and order robustness, while question-aligned
attention constraints further improve performance
under unfavorable orders and reduce prediction variation.

\section{Conclusion}
\label{sec:conclusion}

ViDM demonstrates that multi-question visual inference
can deliver more accurate predictions together with
faster and more stable decisions.
Its central idea is to organize computation and readout
around question-specific decision states, bringing multiple
judgments into a single forward pass rather than
constructing an answer sequence token by token.
Question-aligned attention constraints and multi-order
supervision further reduce prediction variation
under question reordering.
Evaluations across benchmarks and backbones validate
this design across visual tasks and model architectures.

As more questions share the same visual input,
removing sequential answer decoding yields larger
efficiency gains.
For video tasks with substantial visual-input costs,
combining parallel decisions with visual-token compression
offers a direction for further reducing end-to-end costs.
ViDM's role extends beyond standalone answering
to adaptive computation allocation:
its confidence estimates support on-demand invocation
of larger models, allowing high-confidence judgments
to be accepted directly while allocating stronger
computation to the remaining questions.
With ViDM providing answers for approximately 75\%
of the questions and Qwen3.5-27B handling the remainder,
this confidence-guided allocation achieves
end-to-end speedups of $2.6\times$ and $2.9\times$
over Qwen3.5-27B alone on GQA and HallusionBench,
respectively, with corresponding accuracy gains
of 1.68 and 0.25 percentage points.

{
    \small
    \bibliographystyle{ieeenat_fullname}
    \bibliography{main}
}




\clearpage
\setcounter{page}{1}
\maketitlesupplementary

\section{Data Construction and Composition}
\label{sec:supp-data}

\subsection{Construction of Multi-Question Cases}
\label{sec:supp-case-construction}

\begin{figure}[t]
\centering
\begin{minipage}[c]{0.37\linewidth}
  \centering
  \IfFileExists{suppl_1.png}{\includegraphics[width=\linewidth]{suppl_1.png}}%
    {\includegraphics[width=\linewidth]{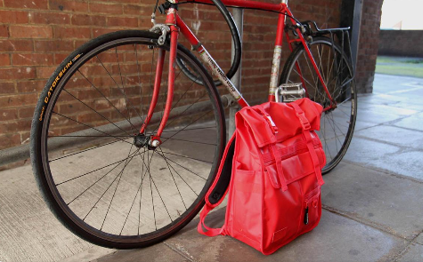}}
\end{minipage}\hfill
\begin{minipage}[c]{0.60\linewidth}
  \small\raggedright
  \textbf{Q1.} What color is the backpack, blue or red? \textbf{Red.}\par\smallskip
  \textbf{Q2.} How many bicycles are visible, one or two? \textbf{One.}\par\smallskip
  \textbf{Q3.} Is the backpack in front of the bicycle, yes or no? \textbf{Yes.}
\end{minipage}
\caption{\textbf{A multi-question case.} The bicycle image from the
framework figure illustrates three decisions over shared visual evidence.
Each question is paired with its own answer (bold).}
\label{fig:supp-case-example}
\end{figure}

We group annotations that refer to the same image or video into a case.
Within each case, we merge repeated questions with matching answers
and discard questions with conflicting labels. Each retained question
keeps its own answer label. Figure~\ref{fig:supp-case-example} shows
how color, count, and spatial judgments can share one image.

\subsection{Training Data Mixture}
\label{sec:supp-training-mixture}

The main training set contains 57,857 distinct images and videos with
158,488 unique media--question pairs. VQAv2 and OCR-VQA together
contribute 69.48\% of source-attributed questions. The two SIMS-VSI
tasks contribute a further 27.44\% through questions about direction
and distance in video. POPE supplies object-presence judgments, and
ChartQA contributes questions grounded in charts
(Table~\ref{tab:supp-training-mixture}).

\begin{table}[t]
\centering
\small
\setlength{\tabcolsep}{3pt}
\caption{\textbf{Training data by source.} Cases count distinct images
or videos within each row. Question counts precede the per-case cap
and order augmentation. Shares use all 158,489 source memberships as
the denominator. One question belongs to two sources.}
\label{tab:supp-training-mixture}
\begin{tabular*}{\columnwidth}{@{\extracolsep{\fill}}lrrr@{}}
\toprule
Source & Cases & Questions & Share (\%) \\
\midrule
VQAv2              & 27,976 & 58,924 & 37.18 \\
POPE               &    430 &  4,333 &  2.73 \\
ChartQA            &    502 &    541 &  0.34 \\
OCR-VQA            & 27,063 & 51,195 & 32.30 \\
SIMS-VSI direction &  2,207 & 32,583 & 20.56 \\
SIMS-VSI distance  &  2,206 & 10,913 &  6.89 \\
\midrule
Distinct total     & 57,857 & 158,488 & --- \\
\bottomrule
\end{tabular*}
\end{table}

Each epoch visits every training case once. Cases with $K=1$, $2$,
$3$, or at least $4$ selected questions contribute 1, 2, 6, or 10
distinct question orders, respectively. This yields 158,971 ordered
training views per epoch. Table~\ref{tab:supp-training-k} shows how this
sampling rule affects the distribution of input lengths. Cases with
$K\geq13$ account for 4.17\% of cases and 15.17\% of training views.

\begin{table}[t]
\centering
\small
\setlength{\tabcolsep}{3pt}
\caption{\textbf{Training cases and views by question count.}
$K$ counts the questions selected for one input, capped at 16.
Each distinct ordering of a case counts as one training view.}
\label{tab:supp-training-k}
\begin{tabular*}{\columnwidth}{@{\extracolsep{\fill}}lrrrr@{}}
\toprule
& \multicolumn{2}{c}{Cases} & \multicolumn{2}{c}{Views per epoch} \\
\cmidrule(lr){2-3}\cmidrule(l){4-5}
$K$ & Count & Share (\%) & Count & Share (\%) \\
\midrule
1         & 16,395 & 28.34 & 16,395 & 10.31 \\
2--4      & 36,957 & 63.88 & 97,526 & 61.35 \\
5--8      &  1,482 &  2.56 & 14,820 &  9.32 \\
9--12     &    612 &  1.06 &  6,120 &  3.85 \\
$\geq13$  &  2,411 &  4.17 & 24,110 & 15.17 \\
\midrule
Total     & 57,857 & 100.00 & 158,971 & 100.00 \\
\bottomrule
\end{tabular*}
\end{table}

\begin{table*}[t]
\centering\small
\setlength{\tabcolsep}{3pt}
\caption{\textbf{End-to-end throughput on nine benchmarks} (questions/s).
Task identifiers follow Table~1 in the main paper. Bold marks the highest
throughput in each column. Dataset values and reported averages are shown
to two decimal places.}
\label{tab:supp-e2e-nine}
\begin{tabular*}{\textwidth}{@{\extracolsep{\fill}}l*{10}{r}@{}}
\toprule
Method & IT-1 & IT-2 & IT-3 & IT-4 & IT-5 & IT-6
       & MIT-1 & VT-1 & VT-2 & Avg. \\
\midrule
Qwen3.5-27B       & 3.38 & 1.99 & 2.02 & 2.14 & 2.48 & 2.02 & 2.27 & 0.44 & 2.15 & 2.10 \\
Qwen3.5-9B        & 4.71 & 2.89 & 2.67 & 2.37 & 3.56 & 3.04 & 4.23 & 0.52 & 3.15 & 3.01 \\
Qwen3.5-4B        & 5.38 & 3.88 & 3.65 & 3.86 & 4.41 & 4.14 & 3.36 & 0.48 & 2.91 & 3.56 \\
InternVL3.5-4B    & 10.26 & 6.28 & 5.52 & 6.07 & 6.10 & 5.53 & 5.17 & 0.41 & 2.86 & 5.35 \\
LLaVA-OneVision-7B & 16.28 & 6.50 & 5.47 & 10.79 & 9.22 & 8.23 & 2.05 & 0.51 & 5.10 & 7.13 \\
Visual-Jev (4B)   & 60.31 & 12.37 & 11.88 & 16.10 & 15.72 & 12.58 & 13.36 & 0.54 & 6.86 & 16.64 \\
Valen-4B          & 96.84 & 20.12 & \textbf{21.72} & 28.48 & 30.00 & 24.74 & 20.67 & 0.52 & 6.50 & 27.73 \\
Vev-4B            & 38.50 & 8.20 & 5.71 & 11.20 & 9.78 & 8.45 & 2.07 & 0.56 & 8.41 & 10.32 \\
\midrule
Qwen3.5-4B + ViDM & \textbf{109.42} & \textbf{20.58} & 21.32 & \textbf{29.98}
                  & \textbf{31.26} & \textbf{25.06} & \textbf{21.54}
                  & \textbf{0.58} & \textbf{9.82} & \textbf{29.95} \\
\bottomrule
\end{tabular*}
\end{table*}

\section{Model Implementation and Training}
\label{sec:supp-implementation}

\subsection{Decision-Query Mask Implementation}
\label{sec:supp-mask}

We assign token ownership on the sequence produced by the multimodal
processor. Shared visual and instruction tokens have owner 0.
Question $i$, its options, and its answer cue have owner $i$.
Starting from the causal mask with padding excluded, we restrict only
the query row at $d_i$ to keys owned by 0 or $i$. Thus each readout
can attend directly to the shared input, its question and options,
and the visible part of its own cue.

\begingroup\footnotesize
\begin{verbatim}
# One sequence; true entries permit attention.
A = causal_mask & valid_key_mask
for i, d in numbered_readout_positions:
    A[d, :] &= (owner == 0) | (owner == i)

for layer in decoder:
    if layer.uses_full_attention:
        layer.use_mask(as_additive_mask(A))
    else:
        layer.use_native_mask()
\end{verbatim}
\endgroup

In Qwen3.5-4B, the eight full-attention layers (layers 4, 8, \ldots,
32, numbered from 1) use this mask for all attention heads. All other
query rows in those layers retain causal attention. The interleaved
linear-attention layers follow the backbone's native computation.

\subsection{Prompt Format, Decision Slots, and Answer-Label Readout}
\label{sec:supp-prompt}

ViDM packs the visual input and all numbered questions into the user
turn of the backbone's chat template. The assistant turn contains one
numbered answer cue per question. The example below illustrates the
prompt layout using the three questions and answer choices in
Figure~\ref{fig:supp-case-example}. Bracketed role names identify the
chat turns, and \texttt{<image>} denotes the image supplied to the processor.

\begingroup\footnotesize
\begin{verbatim}
[system]
You are a visual question-answering model.
Inspect the image. For each question in order,
select one of its listed answers. Do not explain.

[user]
<image>
Question 1: What color is the backpack?
Options:
[0] Red
[1] Blue

Question 2: How many bicycles are visible?
Options:
[0] 1
[1] 2

Question 3: Is the backpack in front
of the bicycle?
Options:
[0] Yes
[1] No

[assistant]
Answer 1:<|fim_suffix|>
Answer 2:<|fim_suffix|>
Answer 3:<|fim_suffix|>
\end{verbatim}
\endgroup

For cue $i$, the readout position $d_i$ is the token immediately
before its marker. We gather the final hidden states at all $K$ such
positions in one forward pass. For each candidate label $a$, let $t(a)$
be the vocabulary ID of its single-token answer string. We copy row
$t(a)$ of the pretrained LM head and its bias, if present. In the
training configuration, option indices 0 and 1 map to the Qwen tokenizer entries for
\texttt{\char32 Yes} and \texttt{\char32 No}, respectively, including
the leading space. All slots share these frozen rows. A softmax over
their scores gives the answer distribution for each question.

\subsection{Training Configuration for the Main Experiments}
\label{sec:supp-main-training}

We initialize ViDM from pretrained Qwen3.5-4B and fine-tune the
language and vision parameters using AdamW. We train on two H200 GPUs
with BF16 and scaled dot-product attention. The global batch contains
16 cases per update, with the remaining cases retained in the last
batch. The language and vision learning rates are $10^{-6}$ and
$2\times10^{-7}$, respectively. We use 5\% linear warmup followed by
cosine decay, AdamW coefficients $(0.9,0.95)$, $\epsilon=10^{-8}$,
weight decay 0.01 on eligible weights, and gradient norm clipping at 1.0.

Each epoch visits the 57,857 training cases in shuffled order. For each
case, we sample $M=1,2,6,10$ distinct permutations for $K=1,2,3,\geq4$,
respectively, using seed 42 with case- and epoch-specific draws.
Labels follow their questions under each permutation. We average
cross-entropy first over questions, then over orders within a case,
and finally over cases in the update. The training schedule spans
15 epochs.

\begin{table*}[t]
\centering\small
\setlength{\tabcolsep}{1.5pt}
\caption{\textbf{Accuracy and efficiency of ViDM-guided Qwen3.5-27B inference.}
Accuracy is averaged over cases. Retained and referred shares are
percentages of questions; the call rate is the percentage of cases that
invoke Qwen3.5-27B. Speedup is relative to standalone Qwen3.5-27B on
the same dataset.
Bold marks the highest accuracy per dataset.}
\label{tab:supp-routing-full}
\begin{tabular*}{\textwidth}{@{\extracolsep{\fill}}llrrrrrr@{}}
\toprule
Dataset & Policy & Acc. (\%) & Retained (\%) & Referred (\%) & Call rate (\%)
& E2E (q/s) & Speedup \\
\midrule
GQA & Qwen3.5-27B & 79.46 & 0.00 & 100.00 & 100 & 3.36 & $1.00\times$ \\
 & ViDM & 78.36 & 100.00 & 0.00 & 0 & 129.52 & $38.56\times$ \\
 & R50 & 79.78 & 53.44 & 46.56 & 94 & 6.04 & $1.80\times$ \\
 & R75 & \textbf{81.14} & 74.27 & 25.73 & 88 & 8.69 & $2.59\times$ \\
 & R90 & 79.44 & 90.10 & 9.90 & 63 & 16.67 & $4.96\times$ \\
 & Random (R75) & 79.35 & 74.27 & 25.73 & 88 & 8.61 & $2.56\times$ \\
\midrule
HallusionBench & Qwen3.5-27B & 91.25 & 0.00 & 100.00 & 100 & 1.87 & $1.00\times$ \\
 & ViDM & 88.42 & 100.00 & 0.00 & 0 & 23.61 & $12.61\times$ \\
 & R50 & \textbf{92.58} & 54.86 & 45.14 & 58 & 3.23 & $1.72\times$ \\
 & R75 & 91.50 & 77.82 & 22.18 & 39 & 5.34 & $2.86\times$ \\
 & R90 & 89.58 & 87.55 & 12.45 & 25 & 7.97 & $4.26\times$ \\
 & Random (R75) & 90.22 & 77.82 & 22.18 & 39 & 5.30 & $2.83\times$ \\
\midrule
VideoHallucer & Qwen3.5-27B & 84.50 & 0.00 & 100.00 & 100 & 0.43 & $1.00\times$ \\
 & ViDM & 82.33 & 100.00 & 0.00 & 0 & 0.58 & $1.36\times$ \\
 & R50 & 84.83 & 56.25 & 43.75 & 54 & 0.32 & $0.75\times$ \\
 & R75 & \textbf{85.50} & 65.34 & 34.66 & 46 & 0.33 & $0.77\times$ \\
 & R90 & 83.00 & 80.68 & 19.32 & 28 & 0.42 & $0.99\times$ \\
 & Random (R75) & 85.44 & 65.34 & 34.66 & 46 & 0.33 & $0.77\times$ \\
\bottomrule
\end{tabular*}
\end{table*}

\subsection{Training Configuration for the Ablations}
\label{sec:supp-ablation-training}

We draw 5,000 distinct training cases without replacement, stratified
jointly by source membership, question-count bin, and image/video
modality. Each of the four configurations in
Table~\ref{tab:supp-training-arms} starts independently from the same
pretrained Qwen3.5-4B checkpoint and uses the optimizer, learning rates,
batch size, frozen readout, and case-balanced loss specified above.

\begin{table}[t]
\centering\small
\caption{\textbf{Matched training configurations.} Every arm uses the
same selected questions and the same number of views per case.}
\label{tab:supp-training-arms}
\begin{tabular}{@{}lll@{}}
\toprule
Configuration & Mask & Orders per case \\
\midrule
Parallel baseline  & Off & Canonical $\times M$ \\
Mask only          & On  & Canonical $\times M$ \\
Multi-order only   & Off & $M$ distinct permutations \\
ViDM               & On  & $M$ distinct permutations \\
\bottomrule
\end{tabular}
\end{table}

With the mask disabled, the parallel baseline still computes all $K$
decisions in one pass using native causal attention. Fixed-order arms
repeat the canonical order to match the number of supervised decisions
in the multi-order arms. Each arm trains for 10 epochs (313 updates per
epoch, 3,130 in total). We evaluate the final checkpoint of every arm.

\section{Inference Efficiency}
\label{sec:supp-efficiency}

\subsection{Single-H200 Timing Protocol}

We evaluate one model at a time on a single H200, using BF16 backbone
weights and one visual case per request. Each request contains all
questions associated with the case. We time the original and reversed
question orders separately. A single-question case requires only one
execution, which contributes once to the throughput calculation.

End-to-end (E2E) timing starts with reading the visual input and ends
when the answers are available for scoring. It covers media decoding,
model-specific preprocessing, host-to-device transfer, visual and
language model execution, and answer parsing or decision readout.
Model loading, warmup, and offline scoring are excluded. For the
Qwen comparison, each request processes the media afresh. Native
generation can reuse its KV cache within a request; ViDM reads all
decisions in one forward pass.

For each benchmark, we divide the total number of questions in the
executed requests by their summed E2E time. For VSR, we sum the question
counts and times across its Random and ZeroShot protocols before taking
the ratio. The overall throughput is the arithmetic mean of the nine
benchmark throughputs, giving each benchmark equal weight.

\subsection{Per-Dataset Efficiency on All Nine Benchmarks}

ViDM has the highest E2E throughput on eight of the nine benchmarks
(Table~\ref{tab:supp-e2e-nine}). Relative to native Qwen3.5-4B, its
average throughput rises from 3.56 to 29.95 questions/s ($8.4\times$).
The largest gain occurs on GQA, from 5.38 to 109.42 questions/s
($20.3\times$). The two video benchmarks show different gains:
$1.2\times$ on VideoHallucer and $3.4\times$ on TempCompass.

\section{ViDM-Guided Inference with Qwen3.5-27B}
\label{sec:supp-routing}

\subsection{Routing Performance on Three Benchmarks}

On each benchmark reported in the main paper's routing table
(GQA, HallusionBench, and VideoHallucer), we evaluate all policies on the same 100 cases, retaining
all questions in their original order. ViDM
scores every question in a case. Answers whose maximum candidate
probability meets the threshold are accepted directly. The remaining
questions are sent to Qwen3.5-27B in a single request, preserving their
relative order. We set the threshold separately for each dataset using
ViDM confidences on 30 disjoint calibration cases. R50, R75, and R90
target retention of 50\%, 75\%, and 90\% of answers, respectively.

The two models execute serially on one H200. We measure throughput
as the number of original questions divided by elapsed pipeline time,
including media preparation, transfer, inference, routing, and answer
merging. Cases that invoke Qwen3.5-27B require a second media decode and
preparation pass. Table~\ref{tab:supp-routing-full} shows that R75 improves
on standalone Qwen3.5-27B by 1.68 accuracy points on GQA and 0.25 on
HallusionBench, with respective speedups of $2.59\times$ and $2.86\times$.
On VideoHallucer, R75 gains 1.00 accuracy point while throughput
falls from 0.43 to 0.33 questions/s.

\subsection{Routing Rates and Question Selection}

Retention measures the fraction of final answers supplied by ViDM.
Referral measures the fraction supplied by Qwen3.5-27B. A case incurs
one large-model call whenever at least one question is referred.
At R75, GQA refers 25.73\% of questions across 88 of 100 cases, whereas
HallusionBench refers 22.18\% across 39 cases. The two datasets thus
send similar fractions of questions to the larger model but differ
substantially in how often a case requires a second pass.

To isolate the effect of question selection at R75, we keep the number
of referrals per case fixed and select the referred questions uniformly
at random. We average random-routing accuracy over three fixed seeds
(113, 227, and 349). Confidence-guided routing improves accuracy over random routing
by 1.79 percentage points on GQA, 1.28 on HallusionBench, and 0.06 on VideoHallucer.

\end{document}